\documentclass[11pt]{article}

\usepackage{nemlap3}
\usepackage{pp-attach}

\title{The effect of alternative tree representations on tree bank grammars}
\author{Mark Johnson\thanks{
I would like to thank Chris Manning, whose observation that PCFG parsers
do not accurately reproduce PP attachment preferences in their training
data stimulated this work, as well as Eugene Charniak, Stuart Geman and 
our students at Brown.}  \\
Cognitive and Linguistic Sciences, Box 1978 \\
Brown University \\
Providence, RI 02912, USA \\
{\tt Mark\_Johnson@Brown.edu}}

\begin{document} \bibliographystyle{fullname}

\maketitle

\begin{abstract}
The performance of PCFGs estimated from tree banks is 
sensitive to the particular way in which linguistic constructions are
represented as trees in the tree bank.  This paper presents a
theoretical analysis of the effect of different tree representations
for PP attachment on PCFG models, and introduces a new methodology for
empirically examining such effects using tree transformations.  It
shows that one transformation, which copies the label of a parent node
onto the labels of its children, can improve the performance of a PCFG
model in terms of labelled precision and recall on held out data from
73\% (precision) and 69\% (recall) to 80\% and 79\% respectively.  It
also points out that if only maximum likelihood parses are of
interest then many productions can be ignored, since they are subsumed
by combinations of other productions in the grammar.  In the Penn~II
tree bank grammar, almost 9\% of productions are subsumed in this way.
\end{abstract}

\section{Introduction}
Parsers which are capable of analysing unrestricted text are of
considerable scientific interest, and have technological applications
in areas such as machine translation and information retrieval as
well.  One way to produce such a parser is to extract a grammar from
one of the larger tree bank corpora currently available.

The relative frequency estimator described below provides a simple way
to estimate from a tree bank corpus a Probabalistic Context Free
Grammar (PCFG) that generates Part Of Speech (POS) tags.  Such a PCFG
induced from a sufficiently large corpus typically generates all
possible POS tag strings.  A parsing system can be obtained by using a
parser to find the maximum likelihood parse tree for an input string.
Such parsing systems often perform as well as other broad coverage
parsing systems for predicting tree structure from POS tags
\cite{Charniak96}.  In addition, many more sophisticated parsing models
are elaborations
of such PCFG models, so understanding the properties of PCFGs is
likely to be useful \cite{Charniak97,Collins97}.

It is well-known that natural language exhibits dependencies that
Context Free Grammars (CFGs), and hence PCFGs, cannot describe
\cite{Shieber:ContextFree}.  But as explained below, the independence
assumptions implicit in PCFGs introduce biases in the statistical
model induced from a tree bank even in constructions which are
adequately described by a CFG.  The direction and size of these 
biases depend on factors such as the following:
\begin{itemize}
\item the precise tree structures used in the tree bank, and
\item whether
      the set of well-formed trees according to the linguistic
      model used to assign trees to strings can be described
      with a CFG.
\end{itemize}

This paper explains how such biases can arise, and presents a series
of experiments in which the trees of a tree bank corpus are
systematically transformed to other tree structures to obtain a
grammar used for parsing, and the inverse tree transform is applied to
the structures produced using this grammar before evaluation.  One of
the transformations described here improves the average
labelled precision and recall on held out data from 73\% (precision)
and 69\% (recall) to 80\% and 79\% respectively.
      
\section{Probabalistic Context Free Grammars}
A PCFG is a CFG in which each production $A \rightarrow \alpha$ in the
grammar's set of productions $R$ is associated with an emission
probability $\P(A \rightarrow \alpha)$ that satisfies a normalization
constraint
\[
 \sum_{\alpha : A \rightarrow \alpha \in R} \P(A \rightarrow \alpha) 
 \; = \; 1
\]
and a consistency or tightness constraint not discussed here.  

A PCFG defines a probability distribution over the (finite) 
parse trees generated by the grammar, where the probability
of a tree $\tau$ is given by
\[
\P(\tau) \; = \; 
 \prod_{A \rightarrow \alpha \in R} 
  \P(A \rightarrow \alpha)^{\C_\tau(A \rightarrow \alpha)}
\]
where $\C_\tau(A \rightarrow \alpha)$ is the `count' of the local tree
consisting of a parent node labelled $A$ with a sequence of immediate
children nodes labelled $\alpha$ in $\tau$, or equivalently, the
number of times the production $A \rightarrow \alpha$ is used in
the derivation $\tau$.

The PCFG which assigns maximum likelihood to a tree bank corpus
$\tilde\tau$ is given by the relative frequency estimator.
\[
\Phat_{\tilde\tau}(A \rightarrow \alpha) \; = \;
 {\C_{\tilde\tau}(A \rightarrow \alpha) \over
 { \sum_{\alpha' : A \rightarrow \alpha' \in R} 
   \C_{\tilde\tau}(A \rightarrow \alpha') }}
\]
Here $\C_{\tilde\tau}(A \rightarrow \alpha)$ refers to the `count' of
the local tree in the tree bank, or equivalently, the number of times
the production $A \rightarrow \alpha$ would be used in derivations
of exactly the trees in $\tilde\tau$.

It is practical to induce PCFGs from tree bank corpora and find
maximum likelihood parses for such PCFGs using relatively modest
computing equipment.  All the experiments reported here used the
Penn~II Wall Street Journal (WSJ) corpus, modified as described by
Charniak \cite{Charniak96}, i.e., empty nodes were deleted, and 
all other components of nodes labels except syntactic category were
removed.

Grammar induction or training used the 39,832 trees in the F2--21
sections of the Penn~II WSJ corpus, and testing was performed on the
1,576 sentences of length~40 or less of the F22 section of the corpus.
Parsing was performed using an exhaustive CKY parser that returned a
maximum likelihood parse.  Ties between equally likely parses were
broken randomly; on the tree bank grammar this leads to fluctuations in
labelled precision and recall with a standard deviation of
approximately 0.07\%.

\section{Different tree structure representations of adjunction}
There is considerable variation in the tree structures used in the
linguistic literature to represent various linguistic constructions.
In this paper we focus on variations in the representation of adjunction
constructions, particularly PP adjunction, but similiar variation
occurs in other constructions as well.

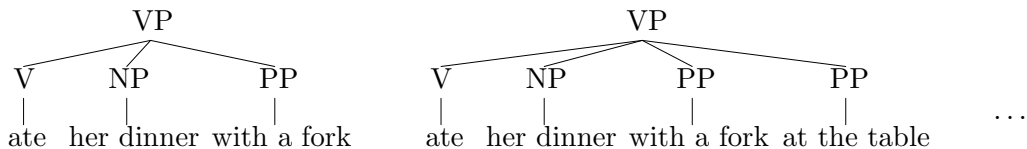
\begin{figure*} 
\begin{center}
\begin{picture}(125,56)(0,-56)
%
\put(47,-8){VP}
\drawline(54,-12)(6,-22)
\put(2,-30){V}
\drawline(6,-34)(6,-44)
\put(0,-52){ate}
\drawline(54,-12)(45,-22)
\put(38,-30){NP}
\drawline(45,-34)(45,-44)
\put(23,-52){her dinner}
\drawline(54,-12)(101,-22)
\put(95,-30){PP}
\drawline(101,-34)(101,-44)
\put(77,-52){with a fork}
\end{picture}
 \hspace{2em} 
\begin{picture}(182,56)(0,-56)
%
\put(76,-8){VP}
\drawline(82,-12)(6,-22)
\put(2,-30){V}
\drawline(6,-34)(6,-44)
\put(-0,-52){ate}
\drawline(82,-12)(45,-22)
\put(38,-30){NP}
\drawline(45,-34)(45,-44)
\put(23,-52){her dinner}
\drawline(82,-12)(101,-22)
\put(95,-30){PP}
\drawline(101,-34)(101,-44)
\put(77,-52){with a fork}
\drawline(82,-12)(159,-22)
\put(153,-30){PP}
\drawline(159,-34)(159,-44)
\put(135,-52){at the table}
\end{picture}
 \hspace{2em}
\raise 0.2in \hbox{\ldots}
\end{center}
\caption{\label{f:flat}
 `Flat' attachment representations of adjunction, where
 adjuncts are attached as siblings of a lexical head (in this
 case, the verb {\em ate}).  
 The Penn~II tree bank represents VP adjunction in this manner.}
\end{figure*}

Early analyses in transformational grammar typically adopted a `flat'
representation of adjunction structures in which adjuncts are
represented as siblings of the phrasal head, as shown in
Figure~\ref{f:flat}.  This representation does not systematically
distinguish between adjuncts and arguments, as both are attached as
children of a single maximal projection. 

The Penn~II tree bank represents PP adjunction to VP in this manner,
presumably because it permits the annotators to avoid having to determine
whether the PP in question is an adjunct or an argument.

Because this representation attaches all of the adjuncts modifying the
same phrase to the same node, distinct CFG productions are required
for each possible number of adjuncts.  Thus the set of all possible
trees following this representation scheme can only be generated by a
CFG if one imposes an upper bound on the number of PPs that can be
adjoined to any one single phrase, but according to standard
linguistic wisdom there is no natural bound on the number of PPs that
may be adjoined to a single phrase.

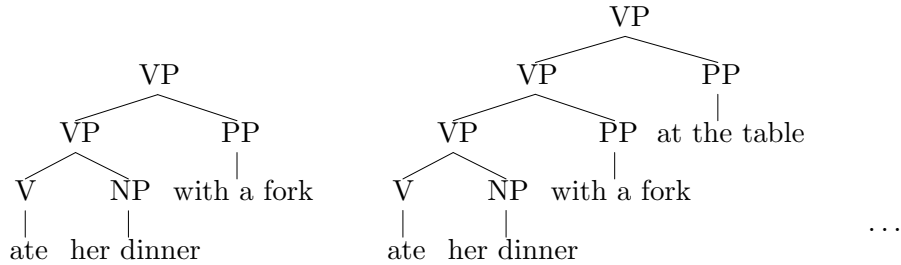
\begin{figure*} 
\begin{center}
\begin{picture}(110,78)(0,-78)
%
\put(49,-8){VP}
\drawline(56,-12)(25,-22)
\put(19,-30){VP}
\drawline(25,-34)(6,-44)
\put(2,-52){V}
\drawline(6,-56)(6,-66)
\put(0,-74){ate}
\drawline(25,-34)(45,-44)
\put(38,-52){NP}
\drawline(45,-56)(45,-66)
\put(23,-74){her dinner}
\drawline(56,-12)(86,-22)
\put(80,-30){PP}
\drawline(86,-34)(86,-44)
\put(62,-52){with a fork}
\end{picture}
 \hspace{2em}
\begin{picture}(149,100)(0,-100)
%
\put(84,-8){VP}
\drawline(90,-12)(56,-22)
\put(49,-30){VP}
\drawline(56,-34)(25,-44)
\put(19,-52){VP}
\drawline(25,-56)(6,-66)
\put(2,-74){V}
\drawline(6,-78)(6,-88)
\put(0,-96){ate}
\drawline(25,-56)(45,-66)
\put(38,-74){NP}
\drawline(45,-78)(45,-88)
\put(23,-96){her dinner}
\drawline(56,-34)(86,-44)
\put(80,-52){PP}
\drawline(86,-56)(86,-66)
\put(62,-74){with a fork}
\drawline(90,-12)(125,-22)
\put(119,-30){PP}
\drawline(125,-34)(125,-44)
\put(102,-52){at the table}
\end{picture}
 \hspace{2em} \raise 0.2in \hbox{\ldots}
\end{center}
\caption{\label{f:chomsky}
 `Chomsky adjunction' representations of adjunction, where each
 adjunct is attached as the unique sibling of a phrasal node (in this
 case, VP). Chomsky's $X'$ theory, used by modern transformational grammar,
 analyses adjunction in a structurally similiar way, except that
 the non-maximal (in these examples, non-root) phrasal nodes are
 given a new category label (in this case ${\rm V}'$).
 }
\end{figure*}

Later transformational analyses adopted the more complex `Chomsky
adjunction' representation of adjunction structures for
theory-internal reasons (e.g., it was a corollary of Emmonds'
``Structure Preserving Hypothesis'').  This representation provides an
additional level of recursive phrasal structure for each adjunct, as
depicted in Figure~\ref{f:chomsky}.

Modern transformational grammar, following Chomsky's $X'$ theory of
phrase structure, represents adjunction with similiar recursive
structures; the major difference being that the non-maximal phrasal
nodes are given a new, distinct category label.

Because the Chomsky adjunction structure and the $X'$ theory based on
it use a single rule to recursively adjoin an arbitrary number of
adjuncts, the set of all tree structures required by this representation
scheme can be generated by a CFG.

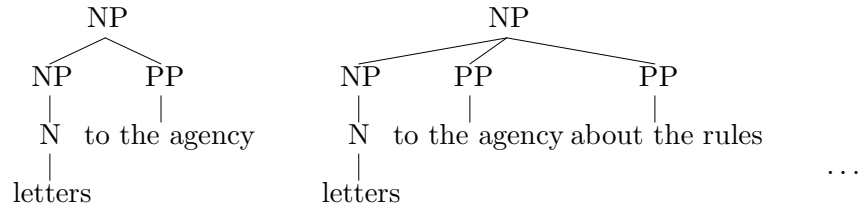
\begin{figure*} 
\begin{center}
\begin{picture}(84,78)(0,-78)
%
\put(28,-8){NP}
\drawline(35,-12)(14,-22)
\put(7,-30){NP}
\drawline(14,-34)(14,-44)
\put(10,-52){N}
\drawline(14,-56)(14,-66)
\put(-0,-74){letters}
\drawline(35,-12)(56,-22)
\put(50,-30){PP}
\drawline(56,-34)(56,-44)
\put(27,-52){to the agency}
\end{picture}
 \hspace{2em}
\begin{picture}(158,78)(0,-78)
%
\put(63,-8){NP}
\drawline(70,-12)(14,-22)
\put(7,-30){NP}
\drawline(14,-34)(14,-44)
\put(10,-52){N}
\drawline(14,-56)(14,-66)
\put(0,-74){letters}
\drawline(70,-12)(56,-22)
\put(50,-30){PP}
\drawline(56,-34)(56,-44)
\put(27,-52){to the agency}
\drawline(70,-12)(126,-22)
\put(120,-30){PP}
\drawline(126,-34)(126,-44)
\put(94,-52){about the rules}
\end{picture}
 \hspace{2em} \raise 0.2in \hbox{\ldots}
\end{center}
\caption{\label{f:penn}
 The representation of NP adjunction used in Penn~II tree bank, where
 adjuncts are attached as siblings of a single NP node.
 }
\end{figure*}

The Penn~II tree bank uses a mixed kind of representation for NP
adjunction, involving two levels of phrasal structure irrespective of
the number of adjuncts, as shown in Figure~\ref{f:penn}.  This
representation permits adjuncts to be systematically distinguished
from arguments, although this does not seem to have been done
systematically in the Penn~II corpus.\footnote{The tree annotation
conventions used in the Penn~II corpus are described in detail in
\cite{Bies95}.  The mixed representation arises from the fact that
``postmodifiers are Chomsky-adjoined to the phrase they modify'' with
the proviso that ``consecutive unrelated adjuncts are non-recursively
attached to the NP the modify''.  However, because constructions such
as appositives, emphatic reflexives and phrasal titles are associated
with their own level of NP structure, it is possible for NPs with more
than two levels of structure to appear.}  Just as with the `flat'
representation, the set of all possible trees required by this mixed
representation cannot be generated by a CFG unless the number of PPs
adjoined to a single phrase is bounded.

Perhaps more seriously for PCFG modelling of such tree structures,
a PCFG which can generate a nontrivial subset of such `two level' NP
tree structures will also generate tree structures which are not
instances of this representational scheme.  For example, the
NP production needed to produce the leftmost tree in Figure~\ref{f:penn}
can apply recursively, generating an alternative tree structure
for the yield of the rightmost tree of Figure~\ref{f:penn},
as shown in Figure~\ref{f:badpenn}.  It is not clear what
interpretation to give tree structures such as these, as they do
not fit the chosen representational scheme for adjunction structures.

\begin{figure}
\begin{center}
\begin{picture}(135,100)(0,-100)
%
\put(62,-8){NP}
\drawline(69,-12)(35,-22)
\put(28,-30){NP}
\drawline(35,-34)(14,-44)
\put(7,-52){NP}
\drawline(14,-56)(14,-66)
\put(10,-74){N}
\drawline(14,-78)(14,-88)
\put(0,-96){letters}
\drawline(35,-34)(56,-44)
\put(50,-52){PP}
\drawline(56,-56)(56,-66)
\put(27,-74){to the agency}
\drawline(69,-12)(103,-22)
\put(98,-30){PP}
\drawline(103,-34)(103,-44)
\put(72,-52){about the rules}
\end{picture}
\end{center}
\caption{\label{f:badpenn}
 A tree structure generated by any PCFG that generates
 the trees in Figure~\ref{f:penn}, yet it does not fit the 
 general representational scheme for adjunction structures used
 in the Penn~II tree bank.}
\end{figure}

\section{PCFG models of PP adjunction} \label{s:theory}

\noindent
This section presents a theoretical investigation into the effect
of different tree representations on the performance of PCFG models
of PP adjunction.  The analysis of four different models is presented
here.

Clearly actual tree bank data is far more complicated than the simple
models investigated in this section, and the next section investigates
the effects of different tree representations empirically by applying
tree transformations to the Penn~II tree bank representations.  However,
the theoretical models discussed in this section show clearly that
the choice of tree representation can in principle affect the
generalizations made by a PCFG model.

\subsection{The Penn~II tree bank representations} \label{s:penn}

\noindent
Suppose we train a PCFG on a corpus $\tilde\tau_1$ consisting only of
two different tree structures: the NP attachment structure labelled
($\A_1$) and the VP attachment tree labelled ($\B_1$). 

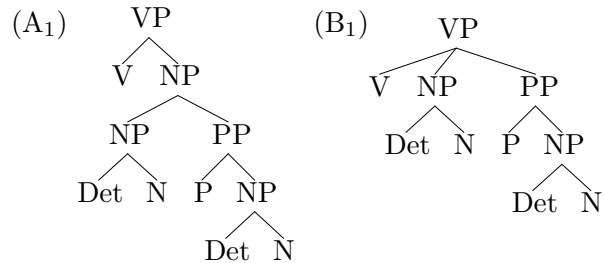
\begin{center} \leavevmode
\raise 1in \hbox{$(\A_1)$} \lower 0.25in \hbox{
\begin{picture}(82,100)(0,-100)
%
\put(20,-8){VP}
\drawline(27,-12)(17,-22)
\put(13,-30){V}
\drawline(27,-12)(38,-22)
\put(31,-30){NP}
\drawline(38,-34)(19,-44)
\put(12,-52){NP}
\drawline(19,-56)(8,-66)
\put(-0,-74){Det}
\drawline(19,-56)(30,-66)
\put(26,-74){N}
\drawline(38,-34)(57,-44)
\put(51,-52){PP}
\drawline(57,-56)(47,-66)
\put(44,-74){P}
\drawline(57,-56)(67,-66)
\put(60,-74){NP}
\drawline(67,-78)(56,-88)
\put(48,-96){Det}
\drawline(67,-78)(78,-88)
\put(74,-96){N}
\end{picture}
}
\raise 1in \hbox{$(\B_1)$}
\begin{picture}(88,78)(0,-78)
%
\put(26,-8){VP}
\drawline(33,-12)(4,-22)
\put(-0,-30){V}
\drawline(33,-12)(25,-22)
\put(18,-30){NP}
\drawline(25,-34)(14,-44)
\put(6,-52){Det}
\drawline(25,-34)(36,-44)
\put(32,-52){N}
\drawline(33,-12)(63,-22)
\put(57,-30){PP}
\drawline(63,-34)(53,-44)
\put(50,-52){P}
\drawline(63,-34)(73,-44)
\put(66,-52){NP}
\drawline(73,-56)(62,-66)
\put(54,-74){Det}
\drawline(73,-56)(84,-66)
\put(80,-74){N}
\end{picture}
\end{center}

In the Penn~II tree bank, structure~($\A_1$) occurs 7,033 times in the
F2-21 subcorpora and 279 times in the F22 subcorpus, and structure~($\B_1$)
occurs 7,717 times in the F2-21 subcorpora and 299 times in the
F22 subcorpus. Thus $f \approx 0.48$ in both the F2-21 subcorpora 
and the F22 corpus.

Returning to the theoretical analysis, 
the relative frequency counts $C_1$ and the non-unit production 
probability estimates $\Phat_1$
for the PCFG induced from this two-tree corpus are as follows:

\begin{center}
\begin{tabular}{l|c|c}
 \multicolumn{1}{c|}{$R$} & $C_1(R)$ & $\Phat_1(R)$ \\ \hline
VP $\rightarrow$ V NP & $f$ & $f$ \\
VP $\rightarrow$ V NP PP & $1-f$ & $1-f$ \\
NP $\rightarrow$ Det N & $2$ & $2/(2+f)$ \\
NP $\rightarrow$ NP PP & $f$ & $f/(2+f)$
\end{tabular}
\end{center}

Of course, in a real tree bank the counts of all these productions
would also include their occurences in other constructions, so the
theoretical analysis presented here is a crude idealization.

Thus the estimated likelihoods using $\Phat_1$ of the tree structures
($\A_1$) and ($\B_1$) are:
\begin{eqnarray*}
\Phat_1(\A_1) & = & { 4 f^2 \over (2+f)^3 } \\
\Phat_1(\B_1) & = & {  4 \, (1 - f) \over (2+f)^2 }.
\end{eqnarray*}

Clearly $\Phat_1(\A_1) < f$ and $\Phat_1(\B_1) < (1-f)$ except at $f=0$ and
$f=1$, so in general the estimated
frequencies using $\Phat_1$ differ from the frequencies of $\A_1$ and $\B_1$
in the training corpus.  This is not too surprising, as the PCFG
$\Phat_1$ assigns non-zero probability to trees not in the training
corpus.  For example, $\Phat_1$ assigns non-zero probability to the
tree in Figure~\ref{f:badpenn}.  We discuss the ramifications of
this in section~\ref{s:subsumedrules}.

In any case, in the parsing applications mentioned earlier the
absolute magnitude of the probability of a tree is not of direct
interest; rather we are concerned with its probability relative to the
probabilities of other, alternative tree structures.  Thus it is
arguably more reasonable to ignore the ``spurious'' tree structures
generated by $\Phat_1$ but not present in the training corpus, and
compare the estimated relative frequencies of ($\A_1$) and ($\B_1$) under
$\Phat_1$ to their frequencies in the training data.

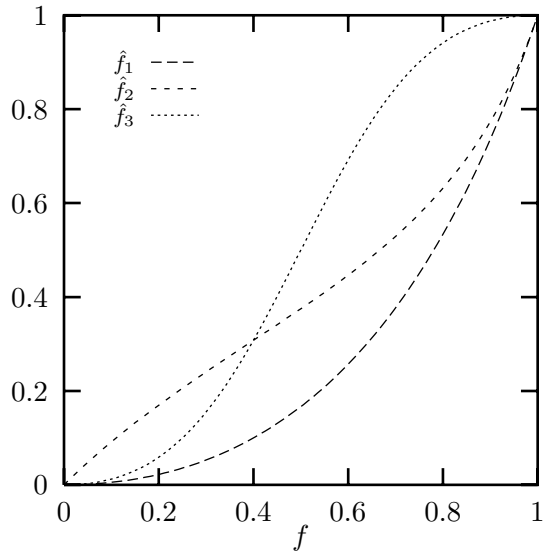
\begin{figure}
\newcommand{\myfhat}{\scriptstyle\fhat}
\hspace{-1.25em}
\setlength{\unitlength}{0.1bp}
\special{!
/gnudict 40 dict def
gnudict begin
/Color false def
/Solid false def
/gnulinewidth 5.000 def
/vshift -33 def
/dl {10 mul} def
/hpt 31.5 def
/vpt 31.5 def
/M {moveto} bind def
/L {lineto} bind def
/R {rmoveto} bind def
/V {rlineto} bind def
/vpt2 vpt 2 mul def
/hpt2 hpt 2 mul def
/Lshow { currentpoint stroke M
  0 vshift R show } def
/Rshow { currentpoint stroke M
  dup stringwidth pop neg vshift R show } def
/Cshow { currentpoint stroke M
  dup stringwidth pop -2 div vshift R show } def
/DL { Color {setrgbcolor Solid {pop []} if 0 setdash }
 {pop pop pop Solid {pop []} if 0 setdash} ifelse } def
/BL { stroke gnulinewidth 2 mul setlinewidth } def
/AL { stroke gnulinewidth 2 div setlinewidth } def
/PL { stroke gnulinewidth setlinewidth } def
/LTb { BL [] 0 0 0 DL } def
/LTa { AL [1 dl 2 dl] 0 setdash 0 0 0 setrgbcolor } def
/LT0 { PL [] 0 1 0 DL } def
/LT1 { PL [4 dl 2 dl] 0 0 1 DL } def
/LT2 { PL [2 dl 3 dl] 1 0 0 DL } def
/LT3 { PL [1 dl 1.5 dl] 1 0 1 DL } def
/LT4 { PL [5 dl 2 dl 1 dl 2 dl] 0 1 1 DL } def
/LT5 { PL [4 dl 3 dl 1 dl 3 dl] 1 1 0 DL } def
/LT6 { PL [2 dl 2 dl 2 dl 4 dl] 0 0 0 DL } def
/LT7 { PL [2 dl 2 dl 2 dl 2 dl 2 dl 4 dl] 1 0.3 0 DL } def
/LT8 { PL [2 dl 2 dl 2 dl 2 dl 2 dl 2 dl 2 dl 4 dl] 0.5 0.5 0.5 DL } def
/P { stroke [] 0 setdash
  currentlinewidth 2 div sub M
  0 currentlinewidth V stroke } def
/D { stroke [] 0 setdash 2 copy vpt add M
  hpt neg vpt neg V hpt vpt neg V
  hpt vpt V hpt neg vpt V closepath stroke
  P } def
/A { stroke [] 0 setdash vpt sub M 0 vpt2 V
  currentpoint stroke M
  hpt neg vpt neg R hpt2 0 V stroke
  } def
/B { stroke [] 0 setdash 2 copy exch hpt sub exch vpt add M
  0 vpt2 neg V hpt2 0 V 0 vpt2 V
  hpt2 neg 0 V closepath stroke
  P } def
/C { stroke [] 0 setdash exch hpt sub exch vpt add M
  hpt2 vpt2 neg V currentpoint stroke M
  hpt2 neg 0 R hpt2 vpt2 V stroke } def
/T { stroke [] 0 setdash 2 copy vpt 1.12 mul add M
  hpt neg vpt -1.62 mul V
  hpt 2 mul 0 V
  hpt neg vpt 1.62 mul V closepath stroke
  P  } def
/S { 2 copy A C} def
end
}
\begin{picture}(2448,2073)(0,0)
\special{"
gnudict begin
gsave
50 50 translate
0.100 0.100 scale
0 setgray
/Helvetica findfont 100 scalefont setfont
newpath
-500.000000 -500.000000 translate
LTa
480 251 M
1785 0 V
480 251 M
0 1771 V
LTb
480 251 M
63 0 V
1722 0 R
-63 0 V
480 605 M
63 0 V
1722 0 R
-63 0 V
480 959 M
63 0 V
1722 0 R
-63 0 V
480 1314 M
63 0 V
1722 0 R
-63 0 V
480 1668 M
63 0 V
1722 0 R
-63 0 V
480 2022 M
63 0 V
1722 0 R
-63 0 V
480 251 M
0 63 V
0 1708 R
0 -63 V
837 251 M
0 63 V
0 1708 R
0 -63 V
1194 251 M
0 63 V
0 1708 R
0 -63 V
1551 251 M
0 63 V
0 1708 R
0 -63 V
1908 251 M
0 63 V
0 1708 R
0 -63 V
2265 251 M
0 63 V
0 1708 R
0 -63 V
480 251 M
1785 0 V
0 1771 V
-1785 0 V
480 251 L
LT0
LT1
808 1845 M
180 0 V
480 251 M
18 0 V
18 0 V
18 1 V
18 0 V
18 1 V
18 1 V
18 2 V
18 1 V
18 2 V
18 2 V
18 2 V
18 2 V
18 2 V
18 3 V
18 3 V
18 3 V
19 4 V
18 3 V
18 4 V
18 4 V
18 5 V
18 4 V
18 5 V
18 5 V
18 6 V
18 5 V
18 6 V
18 6 V
18 7 V
18 7 V
18 7 V
18 7 V
18 8 V
18 8 V
18 8 V
18 9 V
18 9 V
18 9 V
18 10 V
18 10 V
18 11 V
18 10 V
18 11 V
18 12 V
18 12 V
18 12 V
18 13 V
18 13 V
18 13 V
19 14 V
18 15 V
18 14 V
18 16 V
18 15 V
18 16 V
18 17 V
18 17 V
18 18 V
18 18 V
18 19 V
18 19 V
18 20 V
18 20 V
18 21 V
18 21 V
18 22 V
18 23 V
18 23 V
18 24 V
18 25 V
18 25 V
18 26 V
18 27 V
18 27 V
18 28 V
18 29 V
18 30 V
18 30 V
18 31 V
18 32 V
18 33 V
18 34 V
19 35 V
18 35 V
18 37 V
18 37 V
18 39 V
18 39 V
18 41 V
18 42 V
18 42 V
18 44 V
18 46 V
18 46 V
18 47 V
18 49 V
18 51 V
18 51 V
18 53 V
LT2
808 1745 M
180 0 V
480 251 M
18 18 V
18 17 V
18 17 V
18 17 V
18 16 V
18 16 V
18 16 V
18 16 V
18 15 V
18 15 V
18 15 V
18 15 V
18 14 V
18 15 V
18 14 V
18 13 V
19 14 V
18 14 V
18 13 V
18 13 V
18 13 V
18 13 V
18 13 V
18 13 V
18 13 V
18 12 V
18 13 V
18 12 V
18 12 V
18 13 V
18 12 V
18 12 V
18 12 V
18 12 V
18 12 V
18 12 V
18 12 V
18 12 V
18 12 V
18 12 V
18 12 V
18 12 V
18 11 V
18 12 V
18 12 V
18 13 V
18 12 V
18 12 V
18 12 V
19 12 V
18 13 V
18 12 V
18 13 V
18 12 V
18 13 V
18 13 V
18 13 V
18 13 V
18 14 V
18 13 V
18 14 V
18 14 V
18 14 V
18 15 V
18 15 V
18 15 V
18 15 V
18 15 V
18 16 V
18 16 V
18 17 V
18 17 V
18 17 V
18 18 V
18 19 V
18 18 V
18 20 V
18 20 V
18 20 V
18 22 V
18 22 V
18 22 V
19 24 V
18 24 V
18 26 V
18 26 V
18 28 V
18 29 V
18 30 V
18 31 V
18 33 V
18 35 V
18 36 V
18 39 V
18 41 V
18 43 V
18 45 V
18 49 V
18 52 V
LT3
808 1645 M
180 0 V
480 251 M
18 0 V
18 1 V
18 1 V
18 1 V
18 2 V
18 2 V
18 3 V
18 4 V
18 4 V
18 4 V
18 5 V
18 6 V
18 7 V
18 7 V
18 8 V
18 8 V
19 10 V
18 10 V
18 12 V
18 12 V
18 13 V
18 14 V
18 15 V
18 16 V
18 16 V
18 18 V
18 19 V
18 20 V
18 21 V
18 23 V
18 23 V
18 24 V
18 25 V
18 26 V
18 28 V
18 28 V
18 29 V
18 30 V
18 31 V
18 32 V
18 32 V
18 33 V
18 34 V
18 34 V
18 35 V
18 35 V
18 35 V
18 36 V
18 36 V
19 35 V
18 36 V
18 36 V
18 35 V
18 35 V
18 35 V
18 34 V
18 34 V
18 33 V
18 32 V
18 32 V
18 31 V
18 30 V
18 29 V
18 28 V
18 28 V
18 26 V
18 25 V
18 24 V
18 23 V
18 23 V
18 21 V
18 20 V
18 19 V
18 18 V
18 16 V
18 16 V
18 15 V
18 14 V
18 13 V
18 12 V
18 12 V
18 10 V
19 10 V
18 8 V
18 8 V
18 7 V
18 7 V
18 6 V
18 5 V
18 4 V
18 4 V
18 4 V
18 3 V
18 2 V
18 2 V
18 1 V
18 1 V
18 1 V
18 0 V
stroke
grestore
end
showpage
}
\put(748,1645){\makebox(0,0)[r]{$\myfhat_3$}}
\put(748,1745){\makebox(0,0)[r]{$\myfhat_2$}}
\put(748,1845){\makebox(0,0)[r]{$\myfhat_1$}}
\put(1372,51){\makebox(0,0){$f$}}
\put(2265,151){\makebox(0,0){1}}
\put(1908,151){\makebox(0,0){0.8}}
\put(1551,151){\makebox(0,0){0.6}}
\put(1194,151){\makebox(0,0){0.4}}
\put(837,151){\makebox(0,0){0.2}}
\put(480,151){\makebox(0,0){0}}
\put(420,2022){\makebox(0,0)[r]{1}}
\put(420,1668){\makebox(0,0)[r]{0.8}}
\put(420,1314){\makebox(0,0)[r]{0.6}}
\put(420,959){\makebox(0,0)[r]{0.4}}
\put(420,605){\makebox(0,0)[r]{0.2}}
\put(420,251){\makebox(0,0)[r]{0}}
\end{picture}
\caption{\label{f:comparison}
         The estimated normalized frequency $\fhat$ of 
         NP attachment using
         the PCFG models discussed in the text as a function of
         the relative frequency $f$ of NP attachment 
         in the training data.}
\end{figure}

Ideally the estimated
relative frequency $\fhat_1$ of ($\A_1$)
\begin{eqnarray*}
\fhat_1 & = & \Phat_1(\tau = \A_1 : \tau \in \{ \A_1, \B_1 \}) \\
        & = & { \Phat_1(\A_1) \over \Phat_1(\A_1) + \Phat_1(\B_1)}  \\
        & = & { f^2 \over 2-f }
\end{eqnarray*}
will be close to its actual frequency $f$ in the training
corpus.  The relationship between $f$ and $\fhat_1$ is
plotted in Figure~\ref{f:comparison}.  The value of $\fhat_1$ 
can diverge substantially from $f$.  For example, at $f = 0.48$
(the estimate obtained from the Penn~II corpus presented above)
$\fhat_1 = 0.15$.

\subsection{`Chomsky adjunction' representations} \label{s:chomsky}
Now suppose that the corpus contains the following two trees $(\A_2)$
and $(\B_2)$, which are the `Chomsky adjunction' representations
of NP and VP attached PP's respectively, with relative frequencies
$f$ and $(1-f)$ as before.

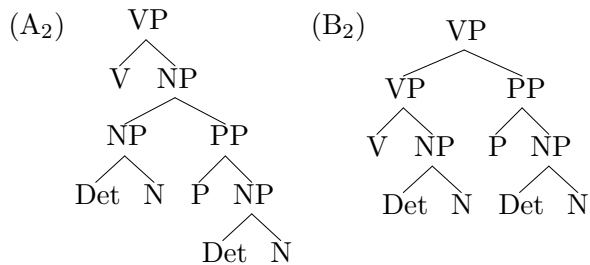
\begin{center} \leavevmode
\raise 1in \hbox{$(\A_2)$} \lower 0.25in \hbox{
\begin{picture}(82,100)(0,-100)
%
\put(20,-8){VP}
\drawline(27,-12)(17,-22)
\put(13,-30){V}
\drawline(27,-12)(38,-22)
\put(31,-30){NP}
\drawline(38,-34)(19,-44)
\put(12,-52){NP}
\drawline(19,-56)(8,-66)
\put(-0,-74){Det}
\drawline(19,-56)(30,-66)
\put(26,-74){N}
\drawline(38,-34)(57,-44)
\put(51,-52){PP}
\drawline(57,-56)(47,-66)
\put(44,-74){P}
\drawline(57,-56)(67,-66)
\put(60,-74){NP}
\drawline(67,-78)(56,-88)
\put(48,-96){Det}
\drawline(67,-78)(78,-88)
\put(74,-96){N}
\end{picture}
}
\raise 1in \hbox{$(\B_2)$}
\begin{picture}(84,78)(0,-78)
%
\put(30,-8){VP}
\drawline(37,-12)(14,-22)
\put(7,-30){VP}
\drawline(14,-34)(4,-44)
\put(0,-52){V}
\drawline(14,-34)(25,-44)
\put(18,-52){NP}
\drawline(25,-56)(14,-66)
\put(6,-74){Det}
\drawline(25,-56)(36,-66)
\put(32,-74){N}
\drawline(37,-12)(59,-22)
\put(53,-30){PP}
\drawline(59,-34)(49,-44)
\put(46,-52){P}
\drawline(59,-34)(69,-44)
\put(62,-52){NP}
\drawline(69,-56)(58,-66)
\put(50,-74){Det}
\drawline(69,-56)(80,-66)
\put(76,-74){N}
\end{picture}
\end{center}

The counts $C_2$ and the non-unit production probability estimates
$\Phat_2$ for the PCFG induced from this two-tree corpus are as
follows:

\begin{center}
\begin{tabular}{l|c|c}
 \multicolumn{1}{c|}{$R$} & $C_2(R)$ & $\Phat_2(R)$ \\ \hline
VP $\rightarrow$ V NP & $1$ & $1/(2-f)$ \\
VP $\rightarrow$ VP PP & $1-f$ & $(1-f)/(2-f)$ \\
NP $\rightarrow$ Det N & $2$ & $2/(2+f)$ \\
NP $\rightarrow$ NP PP & $f$ & $f/(2+f)$
\end{tabular}
\end{center}

The estimated likelihoods using $\Phat_2$ of the tree structures
($\A_2$) and ($\B_2$) are:
\begin{eqnarray*}
\Phat_2(\A_2) & = & {4f \over (4 - f^2)(2 + f)^2} \\
\Phat_2(\B_2) & = & {4 \, (1 - f) \over (4 - f^2)^2}
\end{eqnarray*}

As in the previous subsection $\Phat_2(\A_2) < f$ and $\Phat_2(\B_2) < (1-f)$
because the PCFG assigns non-zero probability to trees not in the
training corpus.  Again, we calculate the estimated relative frequencies
of ($\A_2$) and ($\B_2$) under $\Phat_2$.
\begin{eqnarray*}
\fhat_2 & = & \Phat_2(\tau = \A_2 : \tau \in \{ \A_2, \B_2 \}) \\
        & = & { f^2 - 2f \over 2f^2 - f -2 }
\end{eqnarray*}
The relationship between $f$ and $\fhat_2$ is
plotted in Figure~\ref{f:comparison}.  The value of $\fhat_2$ 
can diverge from $f$, although not as widely as $\fhat_1$.  
For example, at $f = 0.48$
$\fhat_2 = 0.36$.  Thus the precise tree structure representations
used to train a PCFG can have a marked effect on its performance.

\subsection{Penn~II representations with parent annotation} \label{s:parent}
One of the weaknesses of a PCFG is that it is insensitive
to non-local relationships between nodes.  If these relationships
are significant then a PCFG will be a poor language model.
Indeed, the sense in which the set of trees generated by
a CFG is ``context free'' is precisely
that the label on a node completely characterizes the relationships
between the subtree dominated by the node and the set of nodes
that properly dominate this subtree.

Thus one way of relaxing the independence assumptions implicit in a
PCFG model is to systematically encode more information in node
labels about their context.  This subsection explores a particularly
simple kind of contextual encoding: the label of the parent of each non-root
nonpreterminal node is appended to that node's label.
The labels of the root node and the terminal and preterminal nodes
are left unchanged.  

For example, assuming that the Penn~II format trees $(\A_1)$ and
$(\B_1)$ of subsection~\ref{s:penn} are immediately dominated by a
node labelled S, this relabelling applied to those trees produces
the trees $(\A_3)$ and $(\B_3)$ below.

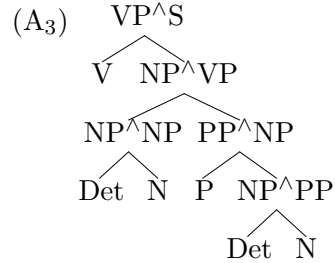
\begin{center} \leavevmode
\raise 1in \hbox{$(\A_3)$} \lower 0.25in \hbox{
\begin{picture}(91,100)(0,-100)
%
\put(12,-8){VP\caret S}
\drawline(25,-12)(9,-22)
\put(5,-30){V}
\drawline(25,-12)(40,-22)
\put(23,-30){NP\caret VP}
\drawline(40,-34)(19,-44)
\put(2,-52){NP\caret NP}
\drawline(19,-56)(8,-66)
\put(0,-74){Det}
\drawline(19,-56)(30,-66)
\put(26,-74){N}
\drawline(40,-34)(61,-44)
\put(45,-52){PP\caret NP}
\drawline(61,-56)(47,-66)
\put(44,-74){P}
\drawline(61,-56)(75,-66)
\put(60,-74){NP\caret PP}
\drawline(75,-78)(64,-88)
\put(56,-96){Det}
\drawline(75,-78)(86,-88)
\put(82,-96){N}
\end{picture}
}
\raise 1in \hbox{$(\B_3)$}
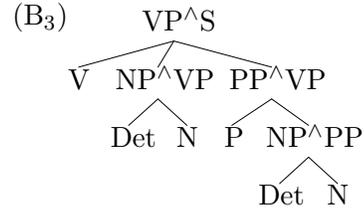
\begin{picture}(106,78)(0,-78)
%
\put(28,-8){VP\caret S}
\drawline(40,-12)(4,-22)
\put(0,-30){V}
\drawline(40,-12)(34,-22)
\put(18,-30){NP\caret VP}
\drawline(34,-34)(23,-44)
\put(16,-52){Det}
\drawline(34,-34)(45,-44)
\put(41,-52){N}
\drawline(40,-12)(77,-22)
\put(61,-30){PP\caret VP}
\drawline(77,-34)(62,-44)
\put(59,-52){P}
\drawline(77,-34)(91,-44)
\put(75,-52){NP\caret PP}
\drawline(91,-56)(80,-66)
\put(72,-74){Det}
\drawline(91,-56)(102,-66)
\put(98,-74){N}
\end{picture}
\end{center}

We can perform the same theoretical analysis on this two tree
corpus that we applied to the previous corpora to investigate
the effect of this relabelling on the PCFG modelling of PP attachment
structures.

The counts $C_3$ and the non-unit production probability estimates
$\Phat_3$ for the PCFG induced from this two-tree corpus are as
follows:

\begin{center}
\begin{tabular}{l|c|c}
 \multicolumn{1}{c|}{$R$} & $C_3(R)$ & $\Phat_3(R)$ \\ \hline
VP\caret S $\rightarrow$ V NP\caret VP & $f$ & $f$ \\
VP\caret S $\rightarrow$ V NP\caret VP PP\caret VP & $1-f$ & $1-f$ \\
NP\caret VP $\rightarrow$ Det N & $1-f$ & $1-f$ \\
NP\caret VP $\rightarrow$ NP\caret NP PP\caret NP & $f$ & $f$
\end{tabular}
\end{center}

The estimated likelihoods using $\Phat_3$ of the tree structures
($\A_3$) and ($\B_3$) are:
\begin{eqnarray*}
\Phat_3(\A_3) & = & f^2 \\
\Phat_3(\B_3) & = & (1-f)^2
\end{eqnarray*}

As in the previous subsection $\Phat_3(\A_3) < f$ and $\Phat_3(\B_3) < (1-f)$.
Again, we calculate the estimated relative frequencies
of ($\A_2$) and ($\B_2$) under $\Phat_2$.
\begin{eqnarray*}
\fhat_3 & = & \Phat_3(\tau = \A_3 : \tau \in \{ \A_3, \B_3 \}) \\
        & = & { f^2 \over f^2 +(1-f)^2 }
\end{eqnarray*}
The relationship between $f$ and $\fhat_3$ is
plotted in Figure~\ref{f:comparison}.  The value of $\fhat_3$ 
can diverge from $f$, just like the other estimates.
For example, at $f = 0.48$
$\fhat_3 = 0.46$.  Thus as expected,
increased context information in the form
of an enriched node labelling scheme can markedly change PCFG
modelling performance.

\section{Tree transformations} \label{s:treetrans}
The last section presented simplified theoretical analyses
of the effect of variation in tree representation and node
labelling on PCFG modelling of PP attachment preferences.
This section reports the results of an empirical investigation
into the effect of changes in tree representation.  These
experiments were conducted by:
\begin{enumerate}
\item systematically transforming the trees in the training corpus F2-21
      by applying a tree transform $X$, 
\item inducing a PCFG $G_X$ from the transformed F2-21 trees, 
\item finding the maximum likelihood parses $Y(\tilde\tau)_X$ of the
      yield of each sentence in the F22 corpus with respect to the PCFG
      $G_X$,
\item applying the inverse transform $X^{-1}$ to these maximum
      likelihood parse trees $Y(\tilde\tau)_X$ to yield a sequence of
      `detransformed' trees $X^{-1}(Y(\tilde\tau)_X)$ using
      (approximately) the same representational system as the tree
      bank itself, and
\item evaluating the detransformed trees  $X^{-1}(Y(\tilde\tau)_X)$
      with the standard labelled precision and recall measures.
\end{enumerate}

Statistics were also collected on the properties of the grammar $G_X$
and its detransformed maximum likelihood parses
$X^{-1}(Y(\tilde\tau)_X)$; the full results are presented in
Table~\ref{t:xf}.

\newcommand{\Q}{?}

\begin{table*}
\begin{center}\begin{tabular}{l*{7}{|c}}
            & \bf F22 &\bf F22 Id&\bf Id&\bf Parent  & \bf VP   &\bf NP   &\bf VP-NP  \\
 \hline
No. of rules    &     & 2,269  & 14,962 & 22,773     & 14,393   & 14,866  & 14,297    \\
Precision       &  1  & 0.772  & 0.735  & 0.801      & 0.722    & 0.738   & 0.730     \\
Recall          &  1  & 0.728  & 0.696  & 0.793      & 0.677    & 0.698   & 0.705     \\
NP attachments  & 279 & 0      & 67     & 217        & 0        & 51      & 329       \\
VP attachments  & 299 & 424    & 384    & 350        & 240      & 427     & 0         \\
NP${}^*$ attachments
                & 339 & 3      & 67     & 234        & 3        & 61      & 401       \\
VP${}^*$ attachments
                & 412 & 668    & 663    & 461        & 493      & 650     & 151
\end{tabular}\end{center}
\caption{\label{t:xf} The results of an empirical study of the effect
 of tree structure on PCFG models.  Each column corresponds to the sequence
 of trees, either consisting of the F22 subcorpus or transforms of the
 maximum likelihood parses of the yields of the F22 subcorpus with
 respect to different PCFGs, as explained in the text.  The first row
 reports the number of productions in these PCFGs, and the next two rows
 give the labelled precision and recall of these sequences of trees.  The
 last four rows report the number of times particular kinds of subtrees
 appear in these sequences of trees, as explained in the text.}
\end{table*}

The columns of that table correspond to different sequences of trees as follows.
\begin{description}
\item{F22:} the trees from the F22 subcorpus of the Penn~II tree bank,
\item{F22 Id:} the maximum likelihood parses of the yields of the F22 subcorpus
 using the PCFG estimated from the F22 subcorpus itself,
\item{Id:} the maximum likelihood parses of the yields of the F22 subcorpus
 using the PCFG estimated from the F2-21 subcorpus (i.e., this corresponds
 to applying an identity transform),
\item{Parent:} as above, except that the parent annotation transform described
 in subsection~\ref{s:parent} was used in training and evaluation, 
\item{VP:} as in {\bf Id}, except that the flat VP structures used
 in the Penn~II tree bank were transformed into recursive Chomsky adjunction structures
 as described below,
\item{NP:} as above, except that the one-level NP structures used in the Penn~II
 tree bank were transformed into recursive Chomsky adjunction structures, and
\item{VP-NP:} as above, except that both NP and VP structures were transformed
 into recursive Chomsky adjunction structures.
\end{description}

The F22 tree sequence column provides information on the distribution of
subtrees in the test tree sequence itself.  The F22 Id PCFG gives data on
the case where the PCFG is trained on the same data that it is
evaluated on, namely the F22 subcorpus.  This column is included
because it is often assumed that the performance of such a model is a
reasonable upper bound on what can be expected from models induced
from training data distinct from the test data.

The remaining columns describe PCFGs induced from versions of the
F2-21 subcorpora obtained by applying tree transformations in the manner
described above.  

The VP transform is the result of exhaustively applying the
tree transforms below.  The first transform transforms VP expansions with final
PPs into Chomsky adjunction structures, and the second transform adjoins final PPs
with a following comma punctuation into Chomsky adjunction structures.
In both cases it is required that the `lowered' sequence of subtrees
$\alpha$ be of length 2 or greater.  This ensures that the transforms
will only apply a finite number of times.  These two rules have the
effect of converting VP final PPs into Chomsky adjunction structures.

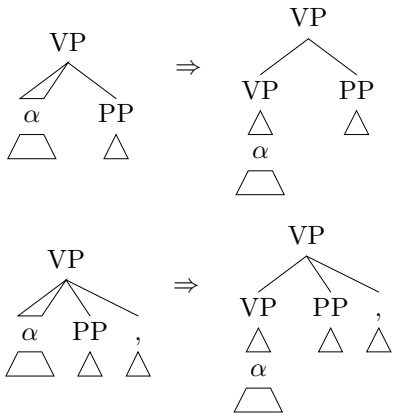
\begin{center}
\setlength{\unitlength}{0.00083333in}
\begingroup\makeatletter\ifx\SetFigFont\undefined%
\gdef\SetFigFont#1#2#3#4#5{%
  \reset@font\fontsize{#1}{#2pt}%
  \fontfamily{#3}\fontseries{#4}\fontshape{#5}%
  \selectfont}%
\fi\endgroup%
{\renewcommand{\dashlinestretch}{30}
\begin{picture}(2387,1161)(0,-10)
\path(2295,762)(1995,987)
\path(1620,162)(1545,12)(1845,12)
	(1770,162)(1620,162)
\path(1695,537)(1620,387)(1770,387)(1695,537)
\path(1995,987)(1695,762)
\path(195,612)(345,612)(495,837)(195,612)
\path(795,612)(495,837)
\path(195,387)(120,237)(420,237)
	(345,387)(195,387)
\path(795,387)(720,237)(870,237)(795,387)
\path(2295,537)(2220,387)(2370,387)(2295,537)
\put(1995,1062){\makebox(0,0)[b]{\smash{{{\SetFigFont{10}{12.0}{\familydefault}{\mddefault}{\updefault}VP}}}}}
\put(1695,237){\makebox(0,0)[b]{\smash{{{\SetFigFont{10}{12.0}{\familydefault}{\mddefault}{\updefault}$\alpha$}}}}}
\put(1695,612){\makebox(0,0)[b]{\smash{{{\SetFigFont{10}{12.0}{\familydefault}{\mddefault}{\updefault}VP}}}}}
\put(1245,762){\makebox(0,0)[b]{\smash{{{\SetFigFont{10}{12.0}{\familydefault}{\mddefault}{\updefault}$\Rightarrow$}}}}}
\put(495,912){\makebox(0,0)[b]{\smash{{{\SetFigFont{10}{12.0}{\familydefault}{\mddefault}{\updefault}VP}}}}}
\put(270,462){\makebox(0,0)[b]{\smash{{{\SetFigFont{10}{12.0}{\familydefault}{\mddefault}{\updefault}$\alpha$}}}}}
\put(795,462){\makebox(0,0)[b]{\smash{{{\SetFigFont{10}{12.0}{\familydefault}{\mddefault}{\updefault}PP}}}}}
\put(2295,612){\makebox(0,0)[b]{\smash{{{\SetFigFont{10}{12.0}{\familydefault}{\mddefault}{\updefault}PP}}}}}
\end{picture}
}
\vspace{1em}
\setlength{\unitlength}{0.00083333in}
\begingroup\makeatletter\ifx\SetFigFont\undefined%
\gdef\SetFigFont#1#2#3#4#5{%
  \reset@font\fontsize{#1}{#2pt}%
  \fontfamily{#3}\fontseries{#4}\fontshape{#5}%
  \selectfont}%
\fi\endgroup%
{\renewcommand{\dashlinestretch}{30}
\begin{picture}(2532,1161)(0,-10)
\path(2145,762)(1995,987)
\path(2445,762)(1995,987)
\path(2145,537)(2070,387)(2220,387)(2145,537)
\path(2445,537)(2370,387)(2520,387)(2445,537)
\path(1620,162)(1545,12)(1845,12)
	(1770,162)(1620,162)
\path(1695,537)(1620,387)(1770,387)(1695,537)
\path(1995,987)(1695,762)
\path(195,612)(345,612)(495,837)(195,612)
\path(645,612)(495,837)
\path(945,612)(495,837)
\path(645,387)(570,237)(720,237)(645,387)
\path(945,387)(870,237)(1020,237)(945,387)
\path(195,387)(120,237)(420,237)
	(345,387)(195,387)
\put(1995,1062){\makebox(0,0)[b]{\smash{{{\SetFigFont{10}{12.0}{\familydefault}{\mddefault}{\updefault}VP}}}}}
\put(2145,612){\makebox(0,0)[b]{\smash{{{\SetFigFont{10}{12.0}{\familydefault}{\mddefault}{\updefault}PP}}}}}
\put(2445,612){\makebox(0,0)[b]{\smash{{{\SetFigFont{10}{12.0}{\familydefault}{\mddefault}{\updefault},}}}}}
\put(1695,237){\makebox(0,0)[b]{\smash{{{\SetFigFont{10}{12.0}{\familydefault}{\mddefault}{\updefault}$\alpha$}}}}}
\put(1695,612){\makebox(0,0)[b]{\smash{{{\SetFigFont{10}{12.0}{\familydefault}{\mddefault}{\updefault}VP}}}}}
\put(1245,762){\makebox(0,0)[b]{\smash{{{\SetFigFont{10}{12.0}{\familydefault}{\mddefault}{\updefault}$\Rightarrow$}}}}}
\put(495,912){\makebox(0,0)[b]{\smash{{{\SetFigFont{10}{12.0}{\familydefault}{\mddefault}{\updefault}VP}}}}}
\put(645,462){\makebox(0,0)[b]{\smash{{{\SetFigFont{10}{12.0}{\familydefault}{\mddefault}{\updefault}PP}}}}}
\put(945,462){\makebox(0,0)[b]{\smash{{{\SetFigFont{10}{12.0}{\familydefault}{\mddefault}{\updefault},}}}}}
\put(270,462){\makebox(0,0)[b]{\smash{{{\SetFigFont{10}{12.0}{\familydefault}{\mddefault}{\updefault}$\alpha$}}}}}
\end{picture}
}
\end{center}

The NP transform is similiar to the VP transform.  It too is the result
of exhaustively applying two tree transformation rules.  These have the
effect of converting NP final PPs into Chomsky adjunction structures.
In this case, we require that $\alpha$ be of length 1 or greater.

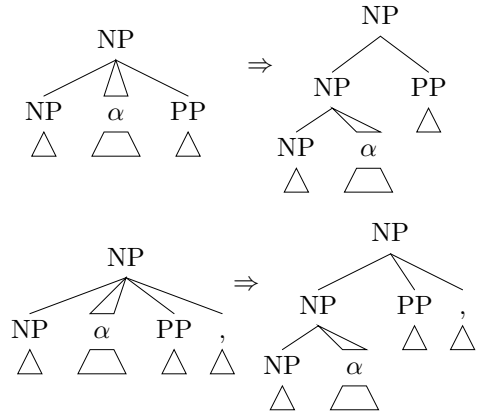
\begin{center}
\setlength{\unitlength}{0.00083333in}
\begingroup\makeatletter\ifx\SetFigFont\undefined%
\gdef\SetFigFont#1#2#3#4#5{%
  \reset@font\fontsize{#1}{#2pt}%
  \fontfamily{#3}\fontseries{#4}\fontshape{#5}%
  \selectfont}%
\fi\endgroup%
{\renewcommand{\dashlinestretch}{30}
\begin{picture}(2600,1161)(0,-10)
\path(2508,762)(2208,987)
\path(2208,987)(1908,762)
\path(1008,612)(558,837)
\path(1008,387)(933,237)(1083,237)(1008,387)
\path(2508,537)(2433,387)(2583,387)(2508,537)
\path(108,387)(33,237)(183,237)(108,387)
\path(483,387)(408,237)(708,237)
	(633,387)(483,387)
\path(483,612)(633,612)(558,837)(483,612)
\path(108,612)(558,837)
\path(2058,162)(1983,12)(2283,12)
	(2208,162)(2058,162)
\path(1908,537)(2058,387)(2208,387)(1908,537)
\path(1683,162)(1608,12)(1758,12)(1683,162)
\path(1683,387)(1908,537)
\put(2208,1062){\makebox(0,0)[b]{\smash{{{\SetFigFont{10}{12.0}{\familydefault}{\mddefault}{\updefault}NP}}}}}
\put(1908,612){\makebox(0,0)[b]{\smash{{{\SetFigFont{10}{12.0}{\familydefault}{\mddefault}{\updefault}NP}}}}}
\put(1458,762){\makebox(0,0)[b]{\smash{{{\SetFigFont{10}{12.0}{\familydefault}{\mddefault}{\updefault}$\Rightarrow$}}}}}
\put(1008,462){\makebox(0,0)[b]{\smash{{{\SetFigFont{10}{12.0}{\familydefault}{\mddefault}{\updefault}PP}}}}}
\put(2508,612){\makebox(0,0)[b]{\smash{{{\SetFigFont{10}{12.0}{\familydefault}{\mddefault}{\updefault}PP}}}}}
\put(108,462){\makebox(0,0)[b]{\smash{{{\SetFigFont{10}{12.0}{\familydefault}{\mddefault}{\updefault}NP}}}}}
\put(558,462){\makebox(0,0)[b]{\smash{{{\SetFigFont{10}{12.0}{\familydefault}{\mddefault}{\updefault}$\alpha$}}}}}
\put(558,912){\makebox(0,0)[b]{\smash{{{\SetFigFont{10}{12.0}{\familydefault}{\mddefault}{\updefault}NP}}}}}
\put(2133,237){\makebox(0,0)[b]{\smash{{{\SetFigFont{10}{12.0}{\familydefault}{\mddefault}{\updefault}$\alpha$}}}}}
\put(1683,237){\makebox(0,0)[b]{\smash{{{\SetFigFont{10}{12.0}{\familydefault}{\mddefault}{\updefault}NP}}}}}
\end{picture}
}
\vspace{1em}
\setlength{\unitlength}{0.00083333in}
\begingroup\makeatletter\ifx\SetFigFont\undefined%
\gdef\SetFigFont#1#2#3#4#5{%
  \reset@font\fontsize{#1}{#2pt}%
  \fontfamily{#3}\fontseries{#4}\fontshape{#5}%
  \selectfont}%
\fi\endgroup%
{\renewcommand{\dashlinestretch}{30}
\begin{picture}(2895,1161)(0,-10)
\path(2508,762)(2358,987)
\path(2358,987)(1908,762)
\path(1008,612)(708,837)
\path(1008,387)(933,237)(1083,237)(1008,387)
\path(2508,537)(2433,387)(2583,387)(2508,537)
\path(108,387)(33,237)(183,237)(108,387)
\path(483,387)(408,237)(708,237)
	(633,387)(483,387)
\path(483,612)(633,612)(708,837)(483,612)
\path(108,612)(708,837)
\path(2058,162)(1983,12)(2283,12)
	(2208,162)(2058,162)
\path(1908,537)(2058,387)(2208,387)(1908,537)
\path(1683,162)(1608,12)(1758,12)(1683,162)
\path(1683,387)(1908,537)
\path(1308,387)(1233,237)(1383,237)(1308,387)
\path(1308,612)(708,837)(708,837)
\path(2808,537)(2733,387)(2883,387)(2808,537)
\path(2808,762)(2358,987)
\put(1908,612){\makebox(0,0)[b]{\smash{{{\SetFigFont{10}{12.0}{\familydefault}{\mddefault}{\updefault}NP}}}}}
\put(1458,762){\makebox(0,0)[b]{\smash{{{\SetFigFont{10}{12.0}{\familydefault}{\mddefault}{\updefault}$\Rightarrow$}}}}}
\put(1008,462){\makebox(0,0)[b]{\smash{{{\SetFigFont{10}{12.0}{\familydefault}{\mddefault}{\updefault}PP}}}}}
\put(2508,612){\makebox(0,0)[b]{\smash{{{\SetFigFont{10}{12.0}{\familydefault}{\mddefault}{\updefault}PP}}}}}
\put(108,462){\makebox(0,0)[b]{\smash{{{\SetFigFont{10}{12.0}{\familydefault}{\mddefault}{\updefault}NP}}}}}
\put(558,462){\makebox(0,0)[b]{\smash{{{\SetFigFont{10}{12.0}{\familydefault}{\mddefault}{\updefault}$\alpha$}}}}}
\put(2133,237){\makebox(0,0)[b]{\smash{{{\SetFigFont{10}{12.0}{\familydefault}{\mddefault}{\updefault}$\alpha$}}}}}
\put(1683,237){\makebox(0,0)[b]{\smash{{{\SetFigFont{10}{12.0}{\familydefault}{\mddefault}{\updefault}NP}}}}}
\put(1308,462){\makebox(0,0)[b]{\smash{{{\SetFigFont{10}{12.0}{\familydefault}{\mddefault}{\updefault},}}}}}
\put(708,912){\makebox(0,0)[b]{\smash{{{\SetFigFont{10}{12.0}{\familydefault}{\mddefault}{\updefault}NP}}}}}
\put(2808,612){\makebox(0,0)[b]{\smash{{{\SetFigFont{10}{12.0}{\familydefault}{\mddefault}{\updefault},}}}}}
\put(2358,1062){\makebox(0,0)[b]{\smash{{{\SetFigFont{10}{12.0}{\familydefault}{\mddefault}{\updefault}NP}}}}}
\end{picture}
}
\end{center}

The NP-VP transform is the result of applying all four of the above
tree transforms.

The rows of Table~\ref{t:xf} provide descriptions of these tree sequences
(after `untransformation', as described above) and, if appropriate,
the PCFGs that generated them.

The labelled precision and recall figures are obtained by regarding
a sequence of trees $\tilde\tau$ as a multiset or bag $E({\tilde\tau})$ of 
edges, i.e., triples 
$\langle N, l, r \rangle$ where $N$ is a nonterminal label and $l$ and
$r$ are left and right string positions in yield of the entire corpus.
(Root nodes and preterminal nodes are ignored in these edge sets, as
they are given as input to the parser).
Relative to a `test sequence' of trees $\tilde\tau'$ (here the F22 subcorpus)
the labelled precision and recall of a  sequence of trees $\tilde\tau$ 
with the same yield as $\tilde\tau'$ are calculated as follows:
\begin{eqnarray*}
\mbox{Precision}(\tilde\tau) & = & { | E(\tilde\tau) \cap E(\tilde\tau') | \over | E(\tilde\tau) |} \\
\mbox{Recall}(\tilde\tau) & = & { | E(\tilde\tau) \cap E(\tilde\tau') | \over | E(\tilde\tau') |} 
\end{eqnarray*}
(The `$\cap$' operation above refers to multiset intersection).
Precision is the fraction of edges in the tree sequence to be
evaluated which also appear in the test sequence, and recall is the
fraction of edges in the test sequence which also appear in sequence
to be evaluated.

The rows labelled NP attachments and VP attachments provide the number
of times the following tree schema, which represent a single 
PP attachment, match the tree sequence.\footnote{The Penn~II markup
scheme permits a `pseudo-attachment' notation for indicating ambiguous
attachment.  However, this is only used relatively infrequently---the
pseudo-attachment markup only appears 27 times in the entire Penn~II
tree bank---and was ignored here.  Pseudo-attachment structures count
as VP attachment structures here.}  In these schema, V can be
instantiated by any of the verbal preterminal tags used in the Penn~II
corpus.

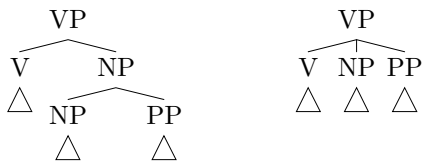
\begin{center}
\setlength{\unitlength}{0.00083333in}
\begingroup\makeatletter\ifx\SetFigFont\undefined%
\gdef\SetFigFont#1#2#3#4#5{%
  \reset@font\fontsize{#1}{#2pt}%
  \fontfamily{#3}\fontseries{#4}\fontshape{#5}%
  \selectfont}%
\fi\endgroup%
{\renewcommand{\dashlinestretch}{30}
\begin{picture}(2579,936)(0,-10)
\path(387,162)(312,12)(462,12)(387,162)
\path(987,162)(912,12)(1062,12)(987,162)
\path(87,462)(12,312)(162,312)(87,462)
\path(87,687)(387,762)(687,687)
\path(387,387)(687,462)(987,387)
\path(1887,687)(2187,762)(2487,687)
\path(2187,762)(2187,687)
\path(1887,462)(1812,312)(1962,312)(1887,462)
\path(2187,462)(2112,312)(2262,312)(2187,462)
\path(2487,462)(2412,312)(2562,312)(2487,462)
\put(387,837){\makebox(0,0)[b]{\smash{{{\SetFigFont{10}{12.0}{\familydefault}{\mddefault}{\updefault}VP}}}}}
\put(87,537){\makebox(0,0)[b]{\smash{{{\SetFigFont{10}{12.0}{\familydefault}{\mddefault}{\updefault}V}}}}}
\put(687,537){\makebox(0,0)[b]{\smash{{{\SetFigFont{10}{12.0}{\familydefault}{\mddefault}{\updefault}NP}}}}}
\put(387,237){\makebox(0,0)[b]{\smash{{{\SetFigFont{10}{12.0}{\familydefault}{\mddefault}{\updefault}NP}}}}}
\put(987,237){\makebox(0,0)[b]{\smash{{{\SetFigFont{10}{12.0}{\familydefault}{\mddefault}{\updefault}PP}}}}}
\put(2187,837){\makebox(0,0)[b]{\smash{{{\SetFigFont{10}{12.0}{\familydefault}{\mddefault}{\updefault}VP}}}}}
\put(1887,537){\makebox(0,0)[b]{\smash{{{\SetFigFont{10}{12.0}{\familydefault}{\mddefault}{\updefault}V}}}}}
\put(2187,537){\makebox(0,0)[b]{\smash{{{\SetFigFont{10}{12.0}{\familydefault}{\mddefault}{\updefault}NP}}}}}
\put(2487,537){\makebox(0,0)[b]{\smash{{{\SetFigFont{10}{12.0}{\familydefault}{\mddefault}{\updefault}PP}}}}}
\end{picture}
}
\end{center}

The rows labelled NP${}^*$ attachments and VP${}^*$ attachments provide
the number of times that the following more relaxed schema match the
tree sequence.  Here $\alpha$ can be instantiated by any sequence of
trees, and V can be instantiated by the same range of preterminal
tags as above.

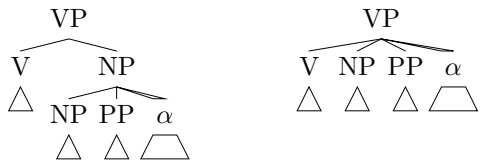
\begin{center}
\setlength{\unitlength}{0.00083333in}
\begingroup\makeatletter\ifx\SetFigFont\undefined%
\gdef\SetFigFont#1#2#3#4#5{%
  \reset@font\fontsize{#1}{#2pt}%
  \fontfamily{#3}\fontseries{#4}\fontshape{#5}%
  \selectfont}%
\fi\endgroup%
{\renewcommand{\dashlinestretch}{30}
\begin{picture}(3056,936)(0,-10)
\path(387,162)(312,12)(462,12)(387,162)
\path(87,462)(12,312)(162,312)(87,462)
\path(87,687)(387,762)(687,687)
\path(387,387)(687,462)(687,387)
\path(1887,687)(2337,762)(2487,687)
\path(2337,762)(2187,687)
\path(1887,462)(1812,312)(1962,312)(1887,462)
\path(2187,462)(2112,312)(2262,312)(2187,462)
\path(2487,462)(2412,312)(2562,312)(2487,462)
\path(687,162)(612,12)(762,12)(687,162)
\path(687,462)(912,387)(987,387)(687,462)
\path(912,162)(837,12)(1137,12)
	(1062,162)(912,162)
\path(2712,462)(2637,312)(2937,312)
	(2862,462)(2712,462)
\path(2337,762)(2712,687)(2787,687)(2337,762)
\put(387,837){\makebox(0,0)[b]{\smash{{{\SetFigFont{10}{12.0}{\familydefault}{\mddefault}{\updefault}VP}}}}}
\put(87,537){\makebox(0,0)[b]{\smash{{{\SetFigFont{10}{12.0}{\familydefault}{\mddefault}{\updefault}V}}}}}
\put(687,537){\makebox(0,0)[b]{\smash{{{\SetFigFont{10}{12.0}{\familydefault}{\mddefault}{\updefault}NP}}}}}
\put(387,237){\makebox(0,0)[b]{\smash{{{\SetFigFont{10}{12.0}{\familydefault}{\mddefault}{\updefault}NP}}}}}
\put(1887,537){\makebox(0,0)[b]{\smash{{{\SetFigFont{10}{12.0}{\familydefault}{\mddefault}{\updefault}V}}}}}
\put(2187,537){\makebox(0,0)[b]{\smash{{{\SetFigFont{10}{12.0}{\familydefault}{\mddefault}{\updefault}NP}}}}}
\put(2487,537){\makebox(0,0)[b]{\smash{{{\SetFigFont{10}{12.0}{\familydefault}{\mddefault}{\updefault}PP}}}}}
\put(687,237){\makebox(0,0)[b]{\smash{{{\SetFigFont{10}{12.0}{\familydefault}{\mddefault}{\updefault}PP}}}}}
\put(987,237){\makebox(0,0)[b]{\smash{{{\SetFigFont{10}{12.0}{\familydefault}{\mddefault}{\updefault}$\alpha$}}}}}
\put(2787,537){\makebox(0,0)[b]{\smash{{{\SetFigFont{10}{12.0}{\familydefault}{\mddefault}{\updefault}$\alpha$}}}}}
\put(2337,837){\makebox(0,0)[b]{\smash{{{\SetFigFont{10}{12.0}{\familydefault}{\mddefault}{\updefault}VP}}}}}
\end{picture}
}
\end{center}

As expected, the PCFG based on the Parent transformation, which
copies the label of each parent node onto those of its children,
outperforms all other PCFGs in terms of labelled precision and recall.

The various adjunction transformations only had minimal effect on 
labelled precision and recall.  Perhaps this is because PP attachment
ambiguities, despite their important role in linguistic and parsing
theory, are just one source of ambiguity among many in real language,
and the effect of the alternative representations has only minor
effect.  

Indeed, in some cases moving to the purportedly linguistically more
realistic tree Chomsky adjunction representations actually decreased
performance on these measures.  On reflection, perhaps this should not
be surprising.  The Chomsky adjunction representations are motivated
within the theoretical framework of Transformational Grammar, which
explicitly argues for nonlocal, indeed, non context free,
dependencies.  Thus its poor performance when used as input to a
statistical model which is insensitive to such dependencies is to
be expected.
Indeed, it might be the case that the additional adjunction nodes
inserted in the tree transformations above have the effect of
converting a local dependency (which can be described by a PCFG)
into a nonlocal dependency (which cannot).

Another initially surprising property of the tree sequences produced
by the PCFGs is that they do not reflect at all well the frequency
of the different kinds of PP attachment found in the Penn~II corpus.
This is in fact to be expected, since the sequences consist of
{\em maximum likelihood} parses.  To see this, consider any of the
examples analysed in section~\ref{s:theory}.
In all of these cases, the corpora contained two tree structures,
and the induced PCFG associates each with an estimated likelihood.
If these likelihoods differ, then a maximum likelihood parser
will always return the same maximum likelihood tree structure
each time it is presented with its yield, and will never return
the tree structure with lower likelihood, even though the
PCFG assigns it a nonzero likelihood.

Thus the surprising fact is that these PCFG parsers ever produce a
nonzero number of NP attachments and VP attachments in the same tree
sequence.  This is possible because the node label V in the attachment
schema above abbreviates several different preterminal labels (i.e.,
the set of all verbal tags).  Further investigation shows that once
the V label in NP and VP attachment schemas is instantiated with a
particular verbal tag, only either the relevant NP attachment schema
or the VP attachment schema appears in the tree sequence.  For instance,
in the Id tree sequence (i.e., produced by the standard tree bank
grammar) the 67 NP attachments all occured with the V label
instantiated to the verbal tag AUX.\footnote{This tag was introduced
by \cite{Charniak96} to distinguish auxiliary verbs from main verbs.}

\section{Subsumed rules in tree bank grammars} \label{s:subsumedrules}
It was mentioned in subsection~\ref{s:penn} that it is
possible for the PCFG induced from a tree bank to generate trees
that are not meaningful representations with respect to the original
tree bank representational scheme.  
The PCFG induced from the F2-21 subcorpus contains the following
two productions:
\begin{eqnarray*}
\Phat(\mbox{NP $\rightarrow$ NP PP}) & = & 0.112 \\
\Phat(\mbox{NP $\rightarrow$ NP PP PP}) & = & 0.006 
\end{eqnarray*}

These productions generate the Penn~II representations of one and two PP 
adjunctions to NP, as explained above.  However, the second of these
productions will never be used in a maximum likelihood parse, as the
parse of the sequence NP PP PP involving two applications of the first
rule has a higher estimated likelihood. 

In fact, {\em all} of the productions of the form $ \mbox{NP}
\rightarrow \mbox{NP}\;\mbox{PP}^n $ where $n>1$ in the PCFG induced
from the F2-21 subcorpus are subsumed by the \mbox{NP $\rightarrow$ NP
PP} production in this way.  Thus PP adjunction to NP in the maximum
likelihood parses using this PCFG always appear as Chomsky
adjunctions, even though the original tree bank did not use this
representational scheme for adjunction!

In fact, a large number of productions in the PCFG induced from the F2-21 
subcorpus are subsumed in this way.  Of the 14,962 productions in
the PCFG, 1,327, or just under 9\%, are subsumed by combinations
of two or more productions.  Since these productions are never
used to construct a maximum likelihood parse, they can be ignored
if only maximum likelihood parses are required.

\section{Conclusion}
There may be several ways of representing a particular linguistic
construction as a tree.  Because of the independence assumptions
implicit in a PCFG, the kind of tree representation employed can
have a dramatic impact on the quality of the PCFG model induced.
This paper introduces a new methodology for examining these
effects utilitizing tree transformations, and showed that one
transformation, which copies the label of a parent node onto the
labels of its children, can dramatically improve the performance
of a PCFG model in terms of labelled precision and recall.  It also
pointed out that if only maximum likelihood parses are of interest
then many productions can be ignored, since they are subsumed
by combinations of other productions in the grammar.

\bibliography{mj}

\begin{thebibliography}{}

\bibitem[\protect\citename{Bies \bgroup et al.\egroup }1995]{Bies95}
Bies, Ann, Mark Ferguson, Karen Katz, and Robert MacIntyre, 1995.
\newblock {\em Bracketting Guideliness for {T}reebank~{II} style {P}enn
  Treebank Project}.
\newblock Linguistic Data Consortium.

\bibitem[\protect\citename{Charniak}1996]{Charniak96}
Charniak, Eugene.
\newblock 1996.
\newblock Tree-bank grammars.
\newblock In {\em Proceedings of the Thirteenth National Conference on
  Artificial Intelligence}, pages 1031--1036, Menlo Park. AAAI Press/MIT Press.

\bibitem[\protect\citename{Charniak}1997]{Charniak97}
Charniak, Eugene.
\newblock 1997.
\newblock Statistical parsing with a context-free grammar and word statistics.
\newblock In {\em Proceedings of the Fourteenth National Conference on
  Artificial Intelligence}, Menlo Park. AAAI Press/MIT Press.

\bibitem[\protect\citename{Collins}1997]{Collins97}
Collins, Michael.
\newblock 1997.
\newblock Three generative, lexicalised models for statistical parsing.
\newblock In {\em The Proceedings of the 35th Annual Meeting of the Association
  for Computational Linguistics}, San Francisco. Morgan Kaufmann.

\bibitem[\protect\citename{Shieber}1985]{Shieber:ContextFree}
Shieber, Stuart~M.
\newblock 1985.
\newblock Evidence against the {C}ontext-{F}reeness of natural language.
\newblock {\em Linguistics and Philosophy}, 8(3):333--344.

\end{thebibliography}

\end{document}